*Original research paper*

# Computer Vision-Aided Intelligent Monitoring of Coffee: Towards Sustainable Coffee Production

Francisco Eron[1]#, Muhammad Noman[1]#, Raphael Ricon de Oliveira[1]#, Deigo de Souza Marques[2], Rafael Serapilha Durelli[2], Andre Pimenta Freire[2], Antonio Chalfun Junior[1]*

[a]Laboratory of Plant Molecular Physiology, Plant Physiology Sector, Department of Biology, Institute of Natural Science Federal University of Lavras, Lavras, MG, Brazil

[b]Department of Computer Science, Federal University of Lavras, Lavras, MG, Brazil

#Equal contribution

*Correspondence: Antonio Chalfun Junior (chalfunjunior@ufla.br)

## Abstract

Coffee which is prepared from the grinded roasted seeds (beans) of harvested coffee cherries, is one of the most consumed beverage and traded commodity, globally. To manually monitor the coffee field regularly, and inform about plant and soil health, as well as estimate yield and harvesting time, is labor-intensive, time-consuming and error-prone. Some recent studies have developed sensors for estimating coffee yield at the time of harvest, however a more inclusive and applicable technology to remotely monitor multiple parameters of the field and estimate coffee yield and quality even at pre-harvest stage, was missing. Following precision agriculture approach, we employed machine learning algorithm YOLO ((You Only Look Once), which works on the basis of Convolutional Neural Network (CNN)), for image processing of coffee plant. In this study, the latest version of the state-of-the-art algorithm YOLOv7 was trained with 324 annotated images followed by its evaluation with 82 unannotated images as test data (supervised method). Next, as an innovative approach for annotating the training data, we trained K-means models which led to machine-generated color classes of coffee fruit and could thus characterize the informed objects in the image (semi-supervised method). Finally, we attempted to develop an AI-based handy mobile application which would not only efficiently predict harvest time, estimate coffee yield and quality, but also inform about plant health. Resultantly, the developed model efficiently analyzed the test data with a mAP@.5 (mean average precision) of 0.89. Strikingly, our innovative semi-supervised method with an m@AP.5 of 0.77 for multi-class mode surpassed the supervised method with m@AP.5 of only 0.60, leading to faster and more accurate annotation. The mobile application we designed based on the developed code, was named "CoffeApp", which possesses multiple features of analyzing fruit from the image taken by phone camera with in field and can thus track fruit ripening in real time. This AI-based technology when integrated with other tools such as UAV would efficiently remotely monitor coffee field for informed decision about irrigation, fertilizer application and other measures of timely field management, hence advancing precision agriculture. Moreover, this machine learning intelligent model could be tailored for various other fruit farming.

## 1. Introduction

Coffee is a highly traded commodity globally, ranking second only to oil in terms of traded value (FAO, 2023). The crop is a major contributor to the socio-economic development of tropical developing countries, with millions of people globally depending on it for their livelihoods. Aside from its contribution to agricultural GDP, coffee production is a significant employer and supports poverty alleviation (Chemura et al., 2016; Läderach et al., 2017). Thus, coffee cultivation is considered an avenue for realizing several of the Sustainable Development Goals (SDGs), such as generating income, creating rural employment, and poverty alleviation (FAO, 2023). Coffee cultivation takes place in over 60 countries, primarily in tropical regions that are conducive to its growth. Brazil, Vietnam, and Colombia are the leading producers worldwide, with Brazil alone accounting for 36% of global coffee production (USDA), while the U.S, Brazil and Europe are its top consumers. Additionally, coffee plantations, especially shaded farms, provide crucial ecosystem services such as biodiversity conservation (Jha et al., 2014), carbon sequestration (van Rikxoort et al., 2014), and soil protection (Meylan et al., 2017).

The coffee market is subject to recurrent supply-demand imbalances and uneven income distribution along the value chain. The global exports of coffee were recorded 10.88 million bags by December 2022 (ICO, 2023). Per data provided by the International Coffee Organization (ICO) in 2023, global coffee production was estimated to have reached 169.34 million bags, with each bag weighing 60 kg, signifying a decline of 2.2% compared to the previous year. In 2021, Brazil suffered a 21.7% drop in coffee production, which amounted to an estimated 67.2 million bags due to weather-associated factors such as drought and frost. The sustainability of coffee bean production and the impact of climate change are key sources of uncertainty for the coffee industry. Climatic conditions, especially during the vegetative and reproductive phases of the coffee plant, significantly influence coffee yield (Tavares et al., 2018). Rising temperatures and precipitation shortages affect flowering, fruiting, and bean quality. Furthermore, climate variability is a key factor influencing the incidence of severe pests and diseases such as coffee leaf rust and coffee berry borer, which can decrease coffee yield and quality and increase production costs (Krishnan, 2017).

Globally, *Coffea arabica* and *Coffea canephora*, commonly referred to as Arabica and Robusta coffees respectively, constitute approximately 99% of the coffee production (Jayakumar et al., 2017). The quality of beans and yield of both species declines when outside these optimal temperature ranges (18-22°C for Arabica, while 22-28°C for Robusta), suggesting significant sensitivity to climatic changes (Magrach and Ghazoul, 2015). Therefore, from a socio-economic standpoint, it is crucial to comprehend the degree of climate-driven impacts on coffee production and the advantages of potential adaptation strategies to maintain and enhance coffee productivity and profitability while sustaining the livelihoods of smallholder producers globally. To protect coffee farms from adverse climatic conditions, keep a sustainable production and even enhance coffee yield and productivity, coffee farms demand continuous monitoring of every aspect.

Nonetheless, the phenomenon of asynchronous flowering poses a significant challenge for coffee growers, leading to irregular fruit ripening (López et al., 2021). This poses problems during the harvesting process, as careful attention must be paid to ensure optimal timing. Oftentimes, for quality coffee production, coffee farmers resort to the practice of lapsed harvesting, wherein they must wait for the next batch of cherries to ripen before harvesting. This approach is not only time-consuming but also labor-intensive, requiring frequent visits and manual screening of the fruits. It is important to note that the quality of coffee is largely dependent on the ripeness of the fruits (Thompson et al., 2012). Coffee fruits, commonly referred to as red cherries, undergo a color transformation during the ripening process (Haile and Kang, 2019). The term "red cherry" is used to describe the fruit's epidermis when it achieves a uniform and intense red color at full ripeness, having progressed through various shades of green, orange, and pink. Overripe cherries turn dark violet, while the presence of green, overripe, or dry cherries in the harvested mass negatively impacts the quality of the beverage and subsequently, its value in the international market (Velásquez et al., 2019). In particular, the proportion of green cherries in the harvested mass can significantly affect the beverage's acidity. To maintain high-quality standards and command a premium price, it is crucial to ensure that at least 98% of the harvested cherries are fully ripe (Leroy et al., 2006).

In the current landscape, the adoption of new technologies and innovation is imperative for the beverage industry to increase productivity and competitiveness. To this end, the scientific community is making significant efforts to develop automatic systems that can enhance the inspection process. Numerous studies have already been conducted, resulting in the development of various applications that have improved for example sorting processes for different fruits and vegetables (Hameed et al., 2018). Technological advancements in precision agriculture play a vital role in obtaining accurate and reliable measurements for crop monitoring. Precision agricultural practices, aimed at achieving high levels of productivity while promoting sustainability, can maximize the potential of each region, resulting in increased crop productivity and quality and reduced cost. Remote sensing has emerged as a promising technology for coffee management, with studies demonstrating its efficacy in evaluating coffee leaf rust levels through the use of Sentinel 2 sensor and Random Forest (RF) algorithms combined with vegetation indices, as described in Chemura et al. research (Chemura et al., 2017).

Computer vision has enabled the implementation of non-destructive techniques for detecting and identifying vegetative structures in the field using images. These techniques have been successfully applied to a wide range of crops including corn (Guerrero et al., 2013) tomatoes (Verma et al., 2014), and oranges (Patel et al., 2011). Besides, these techniques have also been implemented with grapes (Dey et al., 2012), pineapples (Moonrinta et al., 2010), and vegetable crops (Jay et al., 2015). Efficient decision-making on the appropriate harvesting period for coffee fruits can be facilitated by tracking their maturation stages through digital phenotyping. Ramos (2018) suggests that this can be achieved by determining the percentage of mature fruits on tree branches (Ramos et al., 2018). While previous studies have relied on destructive sampling, mainly post-harvest, to quantify and classify fruit for yield estimation (Carrillo and Penaloza, 2009; de Oliveira et al., 2016), only a limited number of studies have explored the classification of coffee fruits before harvest, which can significantly benefit coffee farmers decision-making. Earlier, Avendano et al. (2017)) developed a system that constructs a 3D representation of coffee branches and classifies their vegetative structures (Avendano et al., 2017). In this pursuit, another group came a brilliant idea of developing a CV-based non-destructive method of fruit counting and classification similar to ours (Ramos et al., 2017).

Few advancements were recently seen in this field. For example, a study developed a vegetation index (VI) for coffee ripeness based on the imaging data obtained from coffee fields through an RGB and a five-band multi-spectral cameras, each fixed on a separate UAV (Nogueira Martins et al., 2021). Similarly, Rodriguez et al. earlier came with a classic computer vision approach, however, it involved many instruments for image acquisition, a complex image processing system with precision values (Rodríguez et al., 2020). Although this technique requires the extraction of various features and their input into a classification algorithm, recent advancements in computer vision systems based on deep learning allow for the automatic extraction of multiple features. These techniques have gained popularity due to their speed and accuracy. A very recent study devised yield mapping technique during harvest based on imaging from a camera mounted over the harvesting machine, however, they used an older version of the algorithm, YOLOv4 (Bochkovskiy et al., 2020). Moreover, their mAP value was 83.5% which is lower than ours which is >89% (Bazame et al., 2022). Previously, the same group used a similar approach for mapping coffee yield, however the highest average precision 84% was yet lower (Bazame et al., 2021).

The current study aimed at implementing the state-of-the-art CNN-based computer vision algorithm YOLOv7, the latest version (Wang et al., 2022), to detect and classify coffee fruits on tree branches at different maturation stages. The algorithm was first trained with training data (324 annotated images) and then evaluated with test data (82 unannotated images). We aimed at introducing a novel semi-supervised method for annotating the training data, which would save time and be able to handle large sets of data. Last but not the least, our objective was to develop a handy mobile phone application for efficient image processing. This AI-based technology when integrated with other tools such as UAV would efficiently remotely monitor coffee field for informed decision about irrigation, fertilizer application and other measures of timely field management, implementing precision agriculture in sustainable quality coffee production chain. Nonetheless, this machine learning intelligent model could also be tailored for various other fruit farming.

**2. Materials and Methods**

*2.1. Simple Workflow of developing an ML-based Fruit Detection, Sorting and Quantification*

The first step in developing an AI-based fruit counting and classification system is to collect data on the coffee plants and fruits. This data is usually collected using cameras or sensors that capture images of the plants and fruits, and then annotated. Annotation refers to the process of labeling the data with useful information that can help the machine learning algorithm recognize the different features of the coffee plants and fruits. For instance, in the case of coffee plants, the annotations can include information about the location of the fruits, the number of fruits, their size, shape, and color. Data annotation can be done manually by humans or through automated annotation tools. The next step is to train a machine learning algorithm to recognize and count the fruits, which is done by feeding the algorithm with the annotated images and allowing it to learn to recognize the patterns that distinguish the fruits from other objects in the image. The machine learning algorithm is trained on the training dataset, and its performance is evaluated on the validation dataset. The testing and validation process usually involves comparing the predicted results of the model with the ground truth values of the validation dataset.

*2.2. Data Collection*

Following the approach mentioned above, we began by collecting images from various coffee farms at Lavras, and surrounding areas, Minas Gerais. For a broad applicability, the images used in this study were collected at included images of coffee fruits of every stage as the images were taken along a course of time from unripen green fruits to ripened cherries and raisin ones. The fruit bearing branches were photographed using various smartphone cameras at different angles to achieve representative data.

*2.3. Data Splitting*

The dataset was split into training and validation sets, with a training split of 80% (324 images, 33,717 fruits) and a validation split of 20% (82 images, 10,094 fruits). This random division of the dataset ensures that the models are trained on a diverse set of images and can generalize well to new data.

*2.4. Data Annotation*

The images were manually annotated using Label Studio (Label Studio), an open-source platform for creating labeled datasets, to accurately identify the coffee cherries on the tree canopy. In the process of annotation, the scale presented previously (Ságio, 2009), was used as reference. To facilitate model training, all images were resized to 640 x 640 pixels. To further improve the model ability to generalize, default data augmentation techniques specific to the implemented models were used. In particular, we used mosaic augmentation, as described earlier (Bochkovskiy et al., 2020), to randomly combine multiple images into a single training sample. YOLO is a part of a family of one-stage object detectors and is popular for its speed and accuracy (Wu et al., 2020). Here we also evaluated and compared the efficiency of YOLOv5 (Jocher et al., 2022), 6 (Li et al., 2023) and 7 (Wang et al., 2022), which are the latest and have not been employed before for this purpose. An ideal state-of-the-art model should have (1) a faster and stronger network architecture; (2) a more effective feature integration method; (3) a more accurate detection method; (4) a more robust loss function; (5) a more efficient label assignment method; and (6) a more efficient training method. As compared to YOLOv4, YOLOv7 has been proved to more efficient even with 75% less parameters and 36% less computation (Wang et al., 2022). This approach increased the diversity of the training set and helped the models learn to better handle occlusions and other challenging conditions. Notably, the collected images contained a certain level of noise, which reflects the reality of field data collection and further challenges the models ability to generalize. Equations 1 and 2 were used in comparing the three different models.

$$IoU = \frac{A \cap B}{A \cup B} \tag{1}$$

whereas; IoU - Intersection over Union, A - Ground Truth Boxes, B - Predicted Boxes,

$$APi = \int_0^1 P(R)dR \tag{1}$$

whereas; AP - Average Precision, P – Precision, R – Recall, dR - Recall

*2.5. Developing a semi-supervised annotation system*

To accelerate the annotation process, we utilized the annotation text-file in yolo format (used for training the models in binary and multiclass mode) to collect objects (coffee fruits) in images, which were then subjected to cropping and resizing to 28x28x3 dimensions. To address potential lighting variations, we converted the resized RGB images into the LAB color-space and extracted the A and B color channels for further analysis, eliminating the L channel, since the lightning variation could enhance bias into categorizing fruits based on shadow and light variations. The AB color space images were represented as vectors in a multidimensional space, and K-means models were trained with different k-sizes (2 to 7) on a dataset of approximately 36,000 fruits randomly selected from the dataset to create different classes of colors of fruits.

To enable semi-supervised learning, we curated annotations comprising manually annotated bounding boxes but unsupervised sub-categories of fruits. By leveraging these annotations, we performed semi-supervised learning in object detection tasks in the selected model (Yolov7), which is crucial for real-world applications. By allowing the creation of in-demand complex subcategories of objects, the selected model was trained in the semi-supervised learned sub-categories of fruits and contrasted with the performance from the model of supervised learning from the same number of categories and hyperparameters. The semi-supervised learning categories can be more representative of the mathematical process of categorization in AI and avoid human error being propagated through the machine learning metrics by imposing categories or scales. Besides that, the unsupervised learned categories accelerate the process of annotations and can be used to create mathematically-optimized models and scales.

*2.6. Training Details*

All object detection models were trained using the default hyperparameters specified in the respective papers or repositories, except for the batch size and number of epochs. For this study, a batch size of 16 and 100 epochs were used for all models. The training and evaluation were conducted on a Tesla T4GPU. The evaluation metrics used in this study included Precision (Equation 3), Recall (Equation 4), and mAP (Equation 5).

$$P = \frac{TP}{TP+FP} \tag{3}$$

whereas; P - Precision, TP – Total Positives, FP – False Positives,

$$R = \frac{TP}{TP+FN} \tag{4}$$

whereas; R - Recall, TP – Total Positives, FN - False Negatives

$$mAP@.5 = \frac{1}{n} \sum_{i=0}^{n} AP_i^{0.5} \qquad (5)$$

whereas; mAP - mean Average Precision, AP – Average Precision

*2.7. Validation of the model*

The selected model was used to quantify coffee fruits and their specific class based on their maturity level, such as unripe and ripe. The model was trained on a large dataset of coffee fruit images with labels indicating their class. The model was able to learn features and patterns that distinguish different maturation stages of coffee fruits from each other. The model was then applied to images of the entire dataset of coffee fruits to predict their class and count the number of fruits in each image. The model's performance was evaluated by comparing its predictions with the ground truth labels obtained from the manual annotation process. The model accuracy and precision were reported as metrics of its effectiveness in quantifying coffee fruits and their specific class. The equations 6 and 7 were used to determined ripeness and unripeness.

$$\text{Ripeness (\%)} = \frac{N(ripe\ fruits)}{N(total\ fruits)} \times 100 \qquad (6)$$

$$\text{Unripeness (\%)} = 100 - Ripeness\ (\%) \qquad (7)$$

*2.8. Mobile application development*

Based on the script developed, we designed a mobile phone application, capable of data acquisition and immediate image processing more convenient.

**3. Results and Discussion**

*3.1. Label Studio and Google CoLab are ideal platforms for annotation and Machine learning*

All the images taken in the field were tagged and stored in a local database. To train a computer vision-based machine for example to identify the target object in an image in this case, the algorithm should be first fed with annotated images. In order to train the model for a diverse range of images, we deliberately included images of coffee plant branches in shade and those exposed to sunlight, and even with varied angles and distance. There are several platforms available for image annotation, however, here we annotated our training data through bounding boxes using the online LabelStudio platform. The LabelStudio proved efficient both for aiming the object such as where the object in the image is located as well as classifying it such as what the object is and to which category it belongs. We annotated the whole training data using LabelStudio, while for executing the YOLO through script, we used Google CoLab. At the Google CoLab, the online GPUs allow for faster execution of the code and developing of the model. Google CoLab readily processed all the annotated images and unannotated data. Moreover, it also drew the plots. Figure 1 shows the image before (**1A**) and after (**2B**) annotation with Label Studio. Different colored fruit were labeled with different colored boxes.

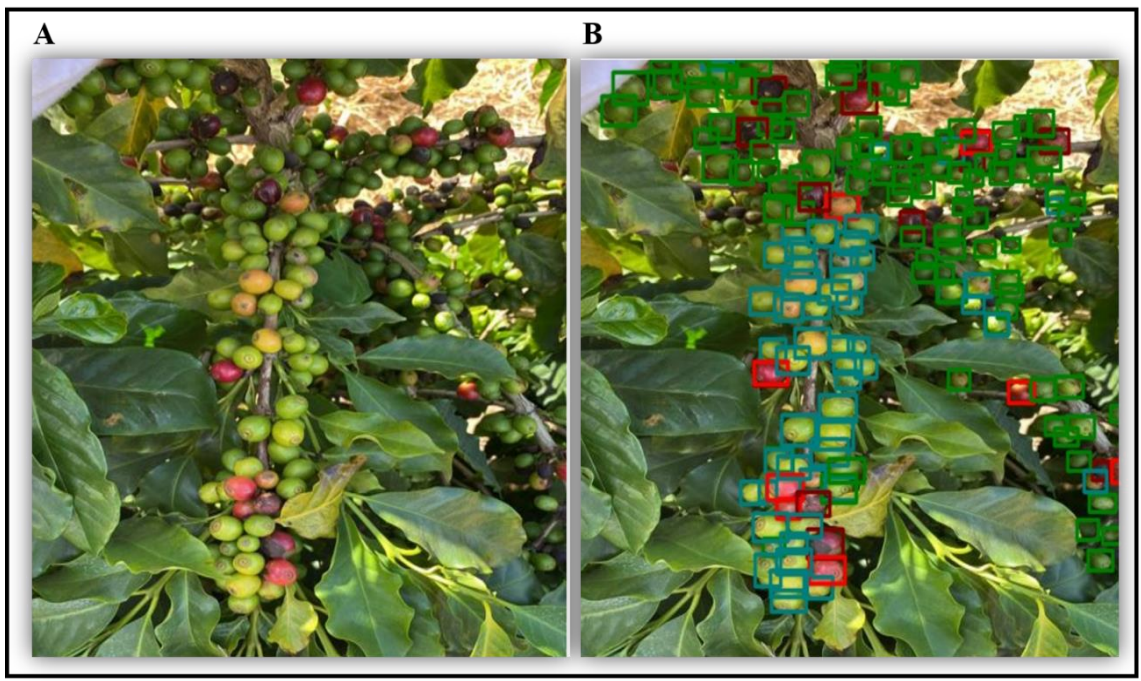

**Figure 1. Annotation of an image using Label Studio.** The unannotated image (**A**) depicts a branch of coffee plant bearing fruits at various stages such as green, yellow, red, raisin and black. Using Label Studio, all the visible fruits in image were annotated (**B**). Different colored boxes were used to classify the fruit at various stages such as unripe, yellow, cherry, raisin and dry. The whole training dataset (324 images) were annotated in the same way.

*3.2. YOLOv7 processed the images with the highest mean average precision*

In the dataset of total 406 images, 324 were used for training data while 82 as test data. After training the algorithm, the selected models (YOLOv5, 6 and 7), chosen for their high mean average precision at 50% intersection over union (mAP@.5) and real-time object detection capabilities in COCO dataset (13; 14), were trained on our training dataset (324 images). While comparing the object detection efficiency of three different YOLO versions, the results showed that YOLOv7 achieved the highest mAP@.5 values in all modes, followed by Yolov5 and Yolov6.. YOLOv7 showed mAP@.5 value of 0.904 for mono class (only fruit), 0.892 for binary class (unripe/ripe) while 0.605 for multiclass (green, yellow-green, cherry, raisin, dry). The rest two versions showed lesser values for all categories (Table 1). This is further elaborated in figure 2 in which all the three versions showed higher performance at mono and binary class but lower at multiclass mode, however, YOLOv7 performed better in all three modes. Though it displayed better results even in multiclass mode, yet it needs further improvement to achieve higher results. Enlarging the training dataset could also further improve its performance.

**Table 1**. **Comparison of object detection performance of three different YOLO versions in three different modes (Mono, Binary, and Multiclass)**. The values of precision (P), recall (R) and mAP@.5 are calculated using the test data. The parameters values indicate the complexity of the models.

| Model | $P$ | $R$ | mAP@.5$_{val}$ | Parameters |
| --- | --- | --- | --- | --- |
| Yolov7(Mono) | 0.852 | 0.871 | **0.904** | 36.9M |
| Yolov7(Binary) | 0.845 | 0.852 | **0.892** | 36.9M |
| Yolov7(Multiclass) | 0.627 | 0.682 | **0.605** | 36.9M |
| Yolov5(Mono) | 0.875 | 0.819 | 0.885 | 21.2M |
| Yolov5(Binary) | 0.844 | 0.821 | 0.866 | 21.2M |
| Yolov5(Multiclass) | 0.64 | 0.562 | 0.555 | 21.2M |
| Yolov6(Mono) | 0.873 | 0.833 | 0.898 | 35.7M |
| Yolov6(Binary) | 0.848 | 0.821 | 0.875 | 35.7M |
| Yolov6(Multiclass) | 0.721 | 0.547 | 0.556 | 35.7M |

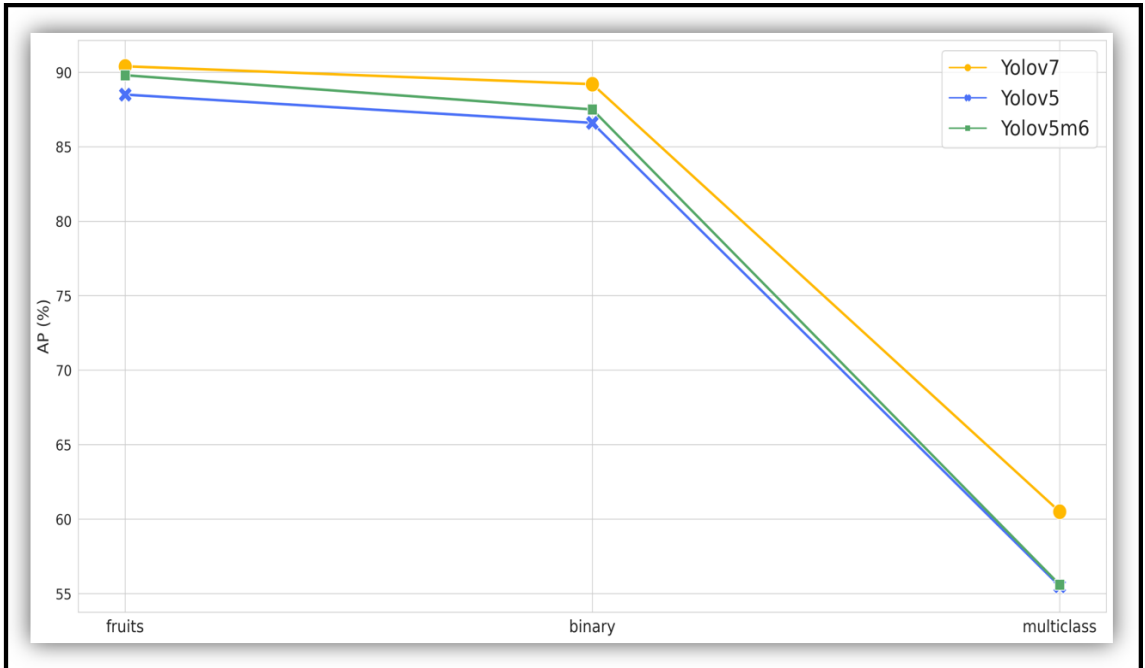

**Figure 2. Comparison of the performance of three different YOLO versions.** For the mean average precision at 50% intersection over union (mAP@.5) in three modes of the dataset: Mono (only fruits), Binary (unripe and ripe fruits), and Multiclass (continuous classification scale - unripe, yellow, cherry, raisin and dry), Yolov7 outperformed Yolv5 and Yolov5m6.

*3.3. A novel semi-supervised annotation system was developed for automatic labelling*

After the model was trained with the training data, and YOLOv7 selected to proceed with, instead of manual annotation (object classification), which is obviously time-consuming and error-prone, we attempted to automate the annotation of images, which was termed as semi-supervised annotation. In this approach, the bounding boxes for training data were yet manual, however, the classification was automatic, hence named as semi-supervised. In other words, we manually located fruits in the image for machine while the machine itself labelled them with their respective categories. This successfully led to the development of a semi-supervised annotation system for training data. Here, K-means clustering was used to create categories of coffee fruits based on their color. We trained K-means models with different k-sizes ranging from 2 to 7 and evaluated their performance based on their ability to identify distinct color clusters. To our interest, the K-means model efficiently identified distinct color clusters within the high-dimensional (28*28*2 axis) and created categories of coffee fruits that were visually distinguishable. In this novel approach, the categories were composed of coffee fruits with similar color representing similar ripening stage, which can be useful for further analysis and classification. After categorization, we performed PCA in the high-dimensional vector to produce visualization of the boundaries among clusters (Figure 3).

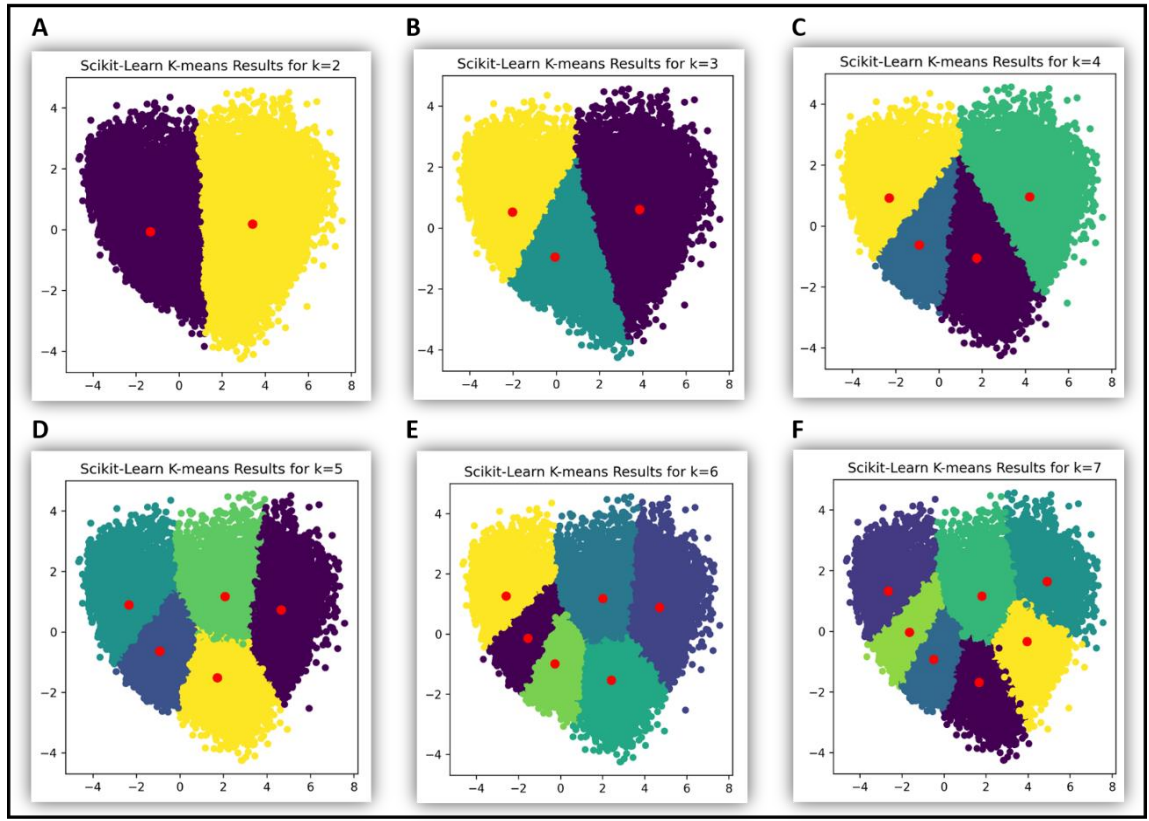

**Figure 3. Principal Component analysis of dataset annotated through semi-supervised approach.** Various clusters in each class are visible in different colors. As the number of classes increased from two to seven classes, the mAP@.5 value also increased, however 4 classes appeared to be optimum.

The seven various color classes of the coffee fruit ripening stage created through k-means are shown in figure 4. Employing semi-supervised approach, we attempted to annotate the same training dataset which was earlier annotated using supervised method. Interestingly, the precision of detection increased with the increase in number of classes, whereas the optimal number of classes was determined to be 4.

However, it is important to note that the categories created by K-means were based solely on color and did not necessarily correspond to different types or varieties of coffee fruits, such as maturity. Therefore, further analysis and classification was required to accurately identify the different types of coffee fruits being represented in each category. The further analysis to acknowledge this problem was performed by correlating the output categories with the Moraes categories used in the annotation process. By doing this, we defined categories in a crescent order of maturity.

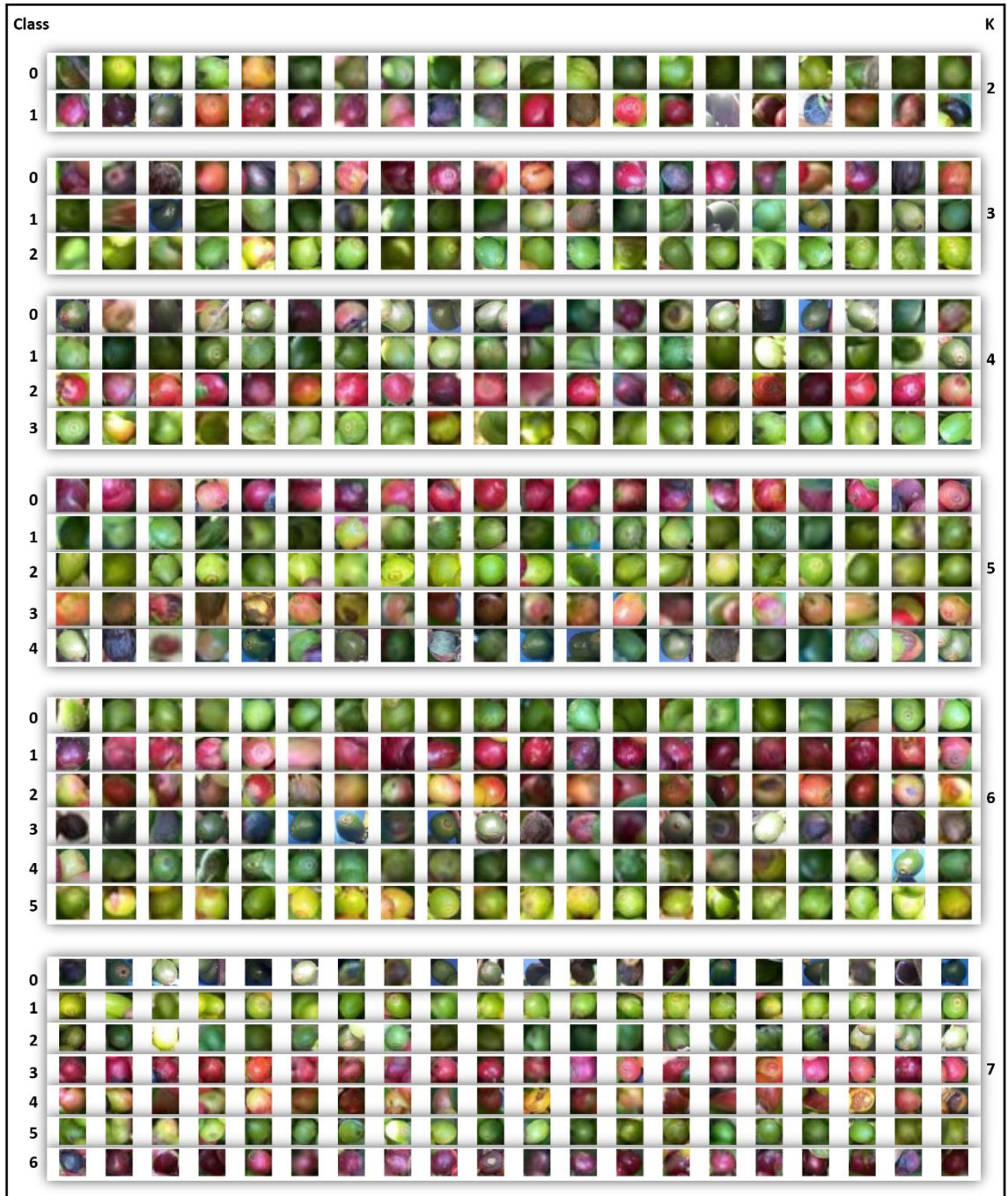

**Figure 3. K-means-based machine–generated color classes as semi-supervised method.** There were up to 7 classes generated keeping k = 7, however the optimal number of classes was determined as 4.

*3.4. Semi-supervised system proved faster and more accurate*

To validate the efficiency as well as consistency of the novel machine learning method, we compared the performance of semi-supervised model with the unsupervised one. Figure 5 shows the comparative performance of supervised and semi-supervised annotations in binary as well as multiclass modes. To our interest, the semi-supervised model performed faster and more accurate annotation than the supervised one. Figure 5A depicts images annotated through supervised (5A) and semi-supervised (5B) method for binary class annotation. The number of ripe (R) and unripe (UR) fruit through ML-based annotation is written above each image. The comparison of supervised and semi-supervised annotation methods in multi-class mode is shown in Figures 5C and 5D, respectively. The number as well as class of the fruit detected are mentioned in the figure. The images in both the binary and multiclass represent four different time points of fruit ripening (from earlier (A, B) to later stages (C, D)). Comparing both the cases (Figure 5A, B and 5C, D), it is clear that semi-supervised annotation surpassed the supervised annotation in terms of speed and accuracy. This is further elaborated graphically in Figure 6. For binary class, the supervised and semi-supervised training models had an equal mAP@.5 of .89 (Figure 6A), showing similar performance for both methods. However for multi-class detection (Figure 6B), the mAP@.5 was 0.77 in case of semi-supervised model, which was only 0.6 with the supervised method, keeping the number of categories the 4 I both cases. It proves the high resolving power of the semi-supervised annotation. Moreover, its faster and more accurate annotation feature will aid in machine learning of large dataset, in less time. This is a novel and rigorous approach to analyze large-scale coffee-fruits datasets, which can have significant implications for various fields such as computer vision, image processing, and machine learning.

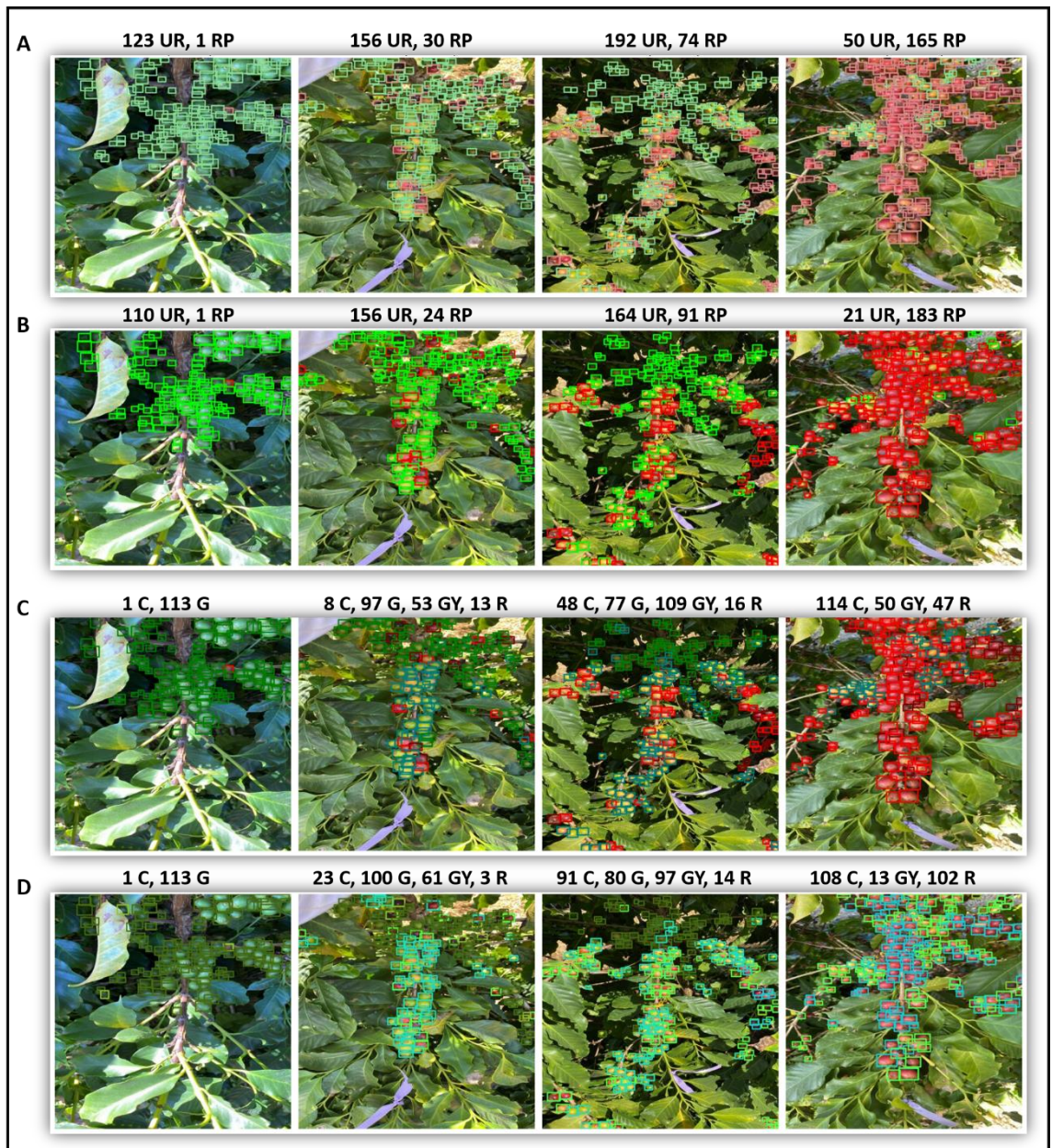

**Figure 5. Comparative performance of supervised and semi-supervised methods.** The novel method of semi-supervised annotation was compared with the supervised for binary (Supervised – **1A** Semi-supervised – **1B**) and multi-class (Supervised – **2A**, Semi-supervised –**2B**) models. The numerals show fruit counts while letters denote fruit type as *UR – Unripe, RP – Ripe, C – Cherry G – Green, GY – Green-yellow, R – Raisin.*

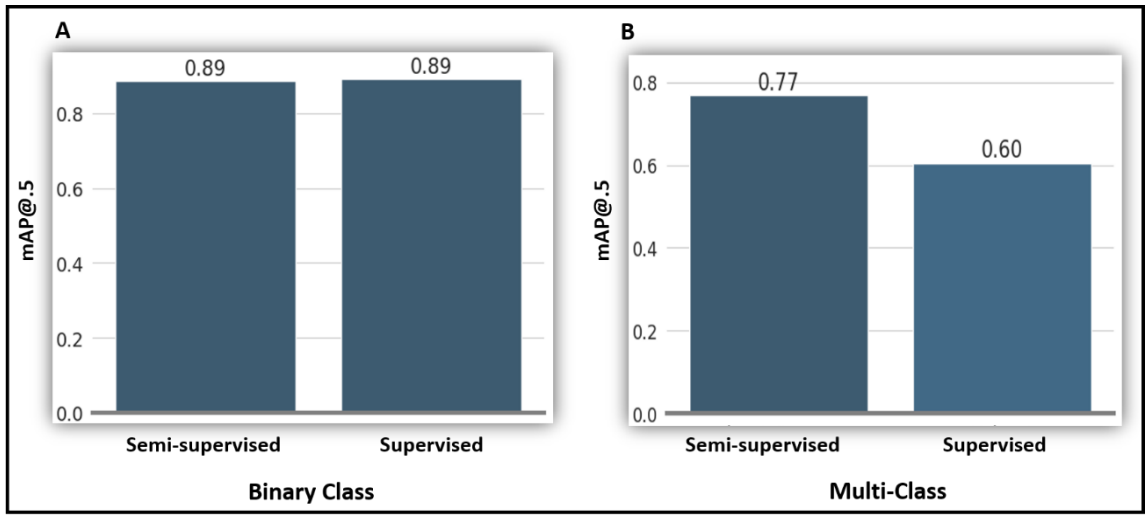

**Figure 6. Graphical representation of the comparative performance of supervised and semi-supervised methods.** For binary class, the supervised and semi-supervised training models had an equal mAP@.5 of .89, showing similar performance for both methods (**A**). However, for multi-class annotation, the semi-supervised method displayed higher mAP@.5 value of 0.77 as compared to 0.66 of the supervised method (**B**), showing better performance.

*3.5. The established model was validated using test images outside our dataset*

To check the efficiency of our trained model, we initially tested it by feeding raw images, not included in our initial dataset. Afterwards, we also tracked the ripening of coffee fruits in real time. Both the approaches proved the image processing efficiency of our established model. Figure 7 depicts raw images not originally included in our dataset. Raw images from the field were analyzed with the model whereas; figure 7A shows the binary class detection counting only ripe and unripe fruits. However, multi-class fruit detection and quantification, classifying them into green, green-yellow, cherry and raisin is also shown in figure 7B. The number and category of the fruit are written above each image. This proved the model was successful in image processing. A collection of data like this will provide a broad picture of the fruit ripening pattern, estimated yield and harvesting time. The big data will eventually aid in informed decision on coffee crop management specially plans for harvest and post-harvest measures.

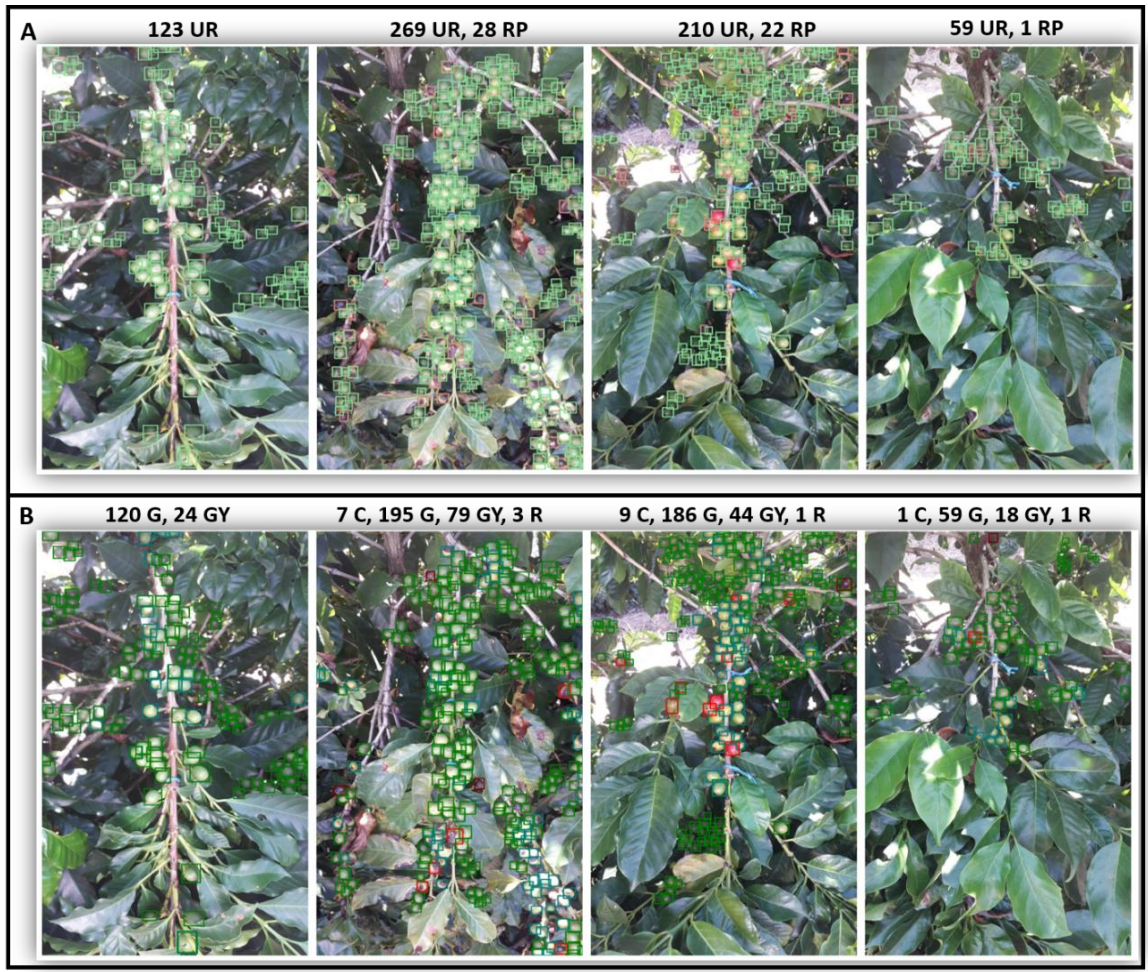

**Figure 7. Validation of the established model using test images outside of dataset.** Raw images from coffee field were analyzed by the for binary (**A**) and multi-class (**B**) fruit detection. The numerals show fruit counts while letters denote fruit type as *UR – Unripe, RP – Ripe, C – Cherry G – Green, GY – Green-yellow, R – Raisin.*

For further validation, we tracked and analyzed the fruit ripening in real time for 90 days. Figure 8A shows the ripening in binary mode (unripe and ripe) over the said time duration. Figure 8B depicts the ripening information of the same data in multi-class mode over the 3-month period. In addition to its yield estimation capabilities, the Yolov7 model can also extract valuable information about the ripening process of crops over time. By calculating the percentage of ripe fruits over months, we can create plots that visualize the progression of ripeness levels as the crops mature, as well as the categorization of the pattern of maturation present in the farm. These plots provide farmers and researchers with valuable insights into the development of the crop, enabling them to plan harvesting schedules, optimize yields, and better understand the underlying biological processes at work. By quantifying ripeness in this way, we can improve our ability to predict and manage crop yields, ultimately leading to more efficient and sustainable agricultural practices. Furthermore, the data collected from the model allowed for an analysis of the distribution of ripe and unripe fruits throughout the growing season.

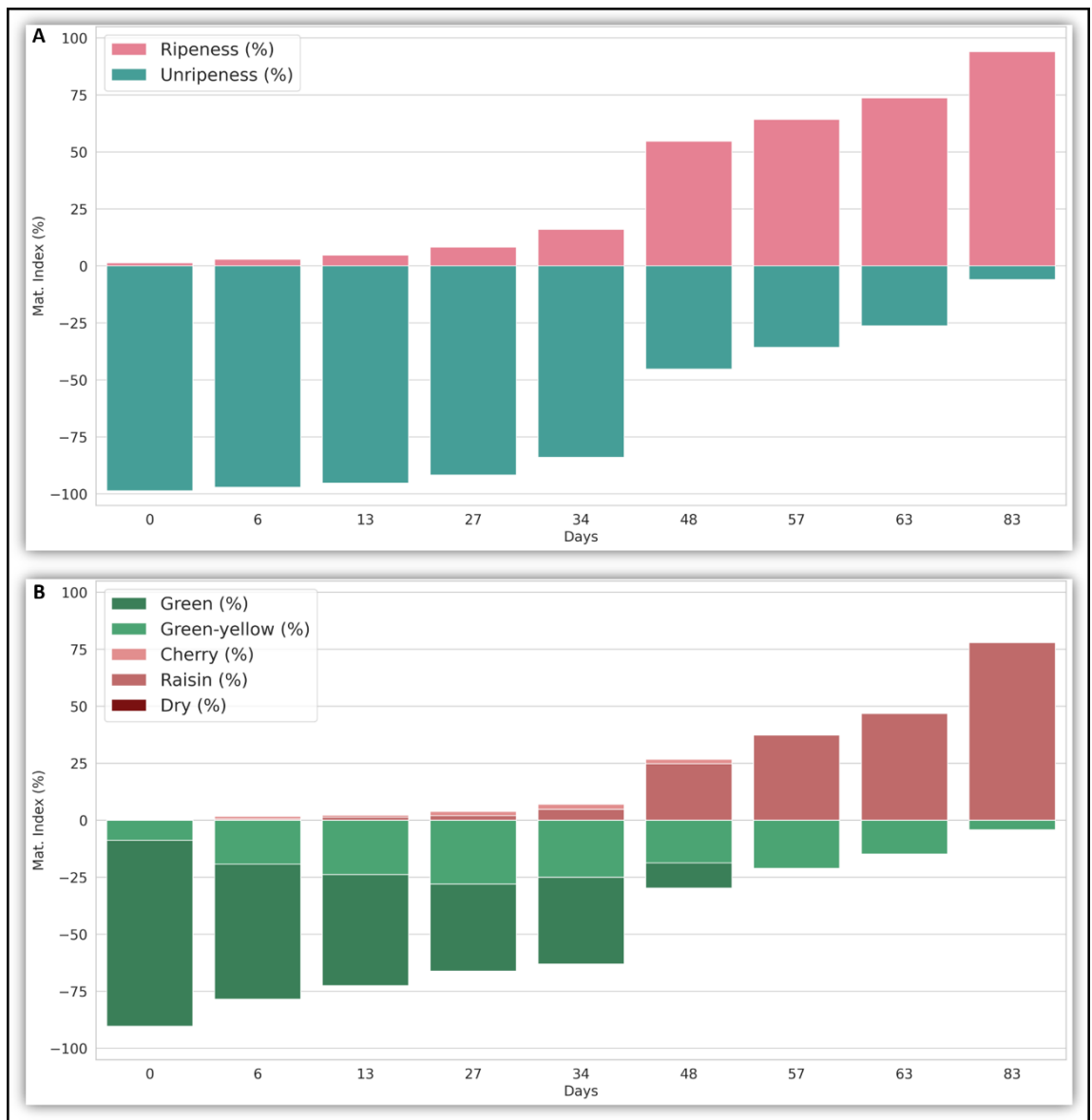

**Figure 8. Tracking coffee fruit ripening with the developed model over a period of 3 months.** Plots show the percentage of ripe and unripe fruits over time in binary (**A**) and multi-class (**B**) modes.

*3.6. CoffeApp: An AI-based multi-functional coffee plant monitoring platform*

The app was developed using the FLutter SDK. Flutter was chosen because it facilitates the generation of the APP for IOS and Android by writing only one source code. Figure 9 presents the architecture behind the APP. Our architecture is based on the micro-service concept. Where we have a backend, developed in Python, using the FastAPI lib. FastAPI framework, high performance, easy to learn, fast to code, ready for production. Note that the App communicates through the HTTP protocol, in other words, a REST API, developed using the Lib FastAPI.

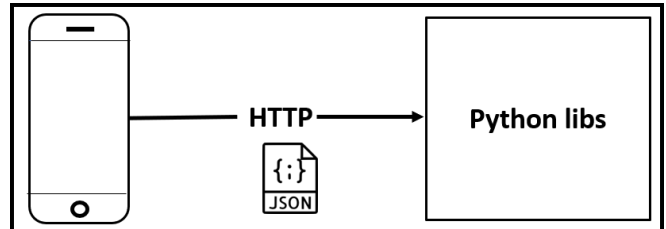

**Figure 9. Architecture of the CoffeApp**.

Various features were incorporated in the application to perform multiple tasks. On the basis of its training, the CoffeApp is able to analyzed diverse array of images. Figure 10 shows glimpses of the graphical user interface of the app, wherein; 10A is the app logo, 10B is the start screen in which multiple options are given to start the analysis with, while 10C shows the parameters such as quantity or class to analyze of coffee fruit, and 10D shows the camera function in the app where branch of coffee fruit is photographed. CoffeApp would aid the farmers and researchers in coffee farming, specially for taking informed decision about plant health, yield estimate, harvest time and post-harvest measures.

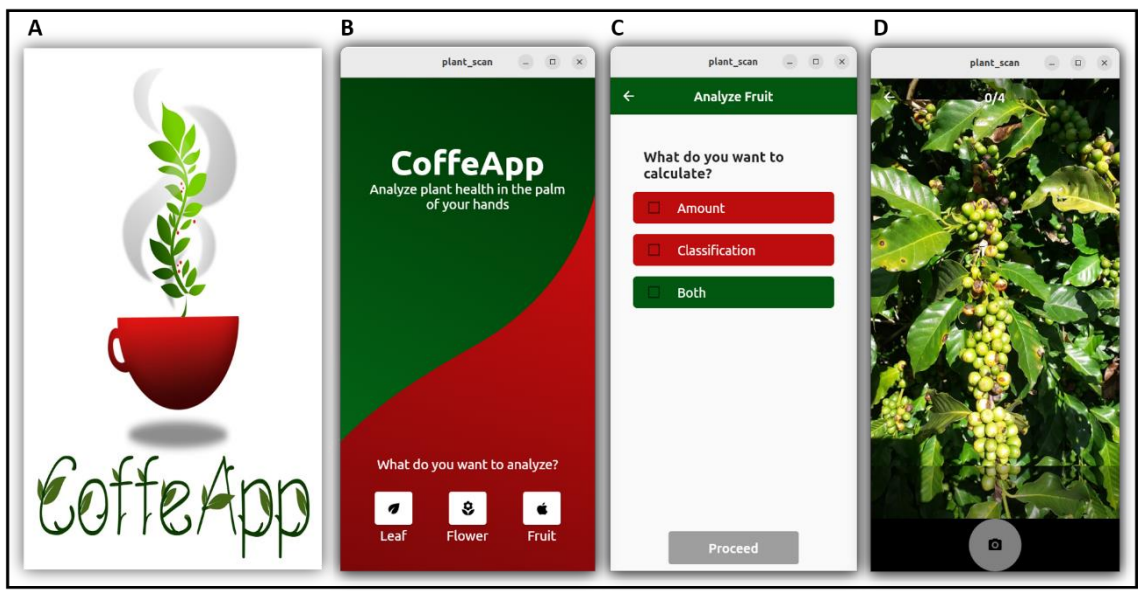

**Figure 10. Graphical user interface (GUI) of the Android application (CoffeApp) for coffee plant monitoring.** The application logo (**A**), home (**B**), fruit analysis parameters such as quantity, class or both (**C**) and camera function to capture images (**D**) are shown.

**Conclusions**

As a temperature sensitive topical plant, weather changes can greatly affect coffee yield. Thus, on-farm non-destructive monitoring of coffee is indispensable for sustainable coffee production. Under these scenarios, computer vision-aided coffee fruit quantification and yield estimation studies have been carried out in the recent past, however some of them used expensive machinery and complex image processing while some used older outdated machine learning models. Moreover, no mobile phone application had been developed for convenient and efficient coffee fruit counting and yield estimation, to date. Using the latest state-of-the-art YOLOv7, we obtained an mAP@.5 of 0.89, the highest ever so far. We also devised a novel method of semi-supervised annotation for the training data. This would greatly aid in handling large datasets and save time. Last but not the least, the CoffeApp, we designed is the first of its kind in CV-aided coffee fruit counting, through which not only researchers but common farmers would estimate the yield as well as harvest time. With the integration of UAV and other value addition, the app holds enormous potential to be used in monitoring coffee farms for informed decision on timely field management, harvest time and post-harvest measures, which will ultimately enhance coffee yield and contribute to sustainable coffee production.

**Acknowledgements**

The authors thank the members of the Laboratory of Plant Molecular Physiology (LFMP, UFLA/Brazil) for structural support of the experiments, and Instituto Brasileiro de Ciência e Tecnologia do Café (INCT-Café) to support plant material and field experiments, Fundaç ão de Amparo à Pesquisa do Estado de Minas Gerais (FAPEMIG, grant number CAP APQ 03605/17), Conselho Nacional de Desenvolvimento Científico e Tecnológico and INCT-Café for the financial support; CNPq for scholarship research for ACJ under grant number 310216/2019-2, and grant #2021/06968-3, São Paulo Research Foundation (FAPESP) for financial support.

**Declaration of competing interest**

The authors declare no competing interest.

**Authors contribution**

**Antonio Chalfun Junior:** Conceptualization, Funding acquisition, Methodology, Project administration Resources, Supervision. **Raphael Ricon de Oliveira:** Conceptualization, Investigation, Methodology, Software, Supervision, Writing - review & editing. **Francisco Eron:** Data curation, Formal analysis, Visualization, Investigation, Software, Writing - original draft. **Muhammad Noman:** Supervision, Visualization, Writing - original draft, Writing - review & editing. **Rafael Serapihla Durelli:** Software. **Deigo de Souza Marque:** Software. **Andre Pimenta Friere:** Software.